# Using Industrial Robots to Manipulate the Measured Object in CMM

Regular Paper

Samir Lemes[1,*], Damir Strbac[1] and Malik Cabaravdic[1]

1 Mechanical Engineering Faculty, University of Zenica, Zenica, Bosnia and Herzegovina
* Corresponding author E-mail: slemes@mf.unze.ba

Abstract Coordinate measuring machines (CMMs) are widely used to check dimensions of manufactured parts, especially in the automotive industry. The major obstacles in automation of these measurements are fixturing and clamping assemblies, which are required in order to position the measured object within the CMM. This paper describes how an industrial robot can be used to manipulate the measured object within the CMM work space, in order to enable automation of complex geometry measurement.

Keywords Industrial Robot, CMM, Measurement

## 1. Introduction

In today's global market, the importance of correct, reliable and comparable measurements is the key factor for achieving quality in activities and procedures in every area of industry. Calibration, testing and measurement are necessary elements in the development process or progress in many disciplines of science and industry.

The majority of modern industrial measurements can be categorized as GDT (Geometrical Dimensioning and Tolerancing). Increasingly, measurements obtained by coordinate measuring machines (CMM) are being used.

CMMs are measuring devices with high measuring speed. Positioning and rotation of the measured object are always performed manually in the work area of the coordinate measuring machine. The object, with defined dimensions, shape and measuring surfaces, is measured and controlled from several different sides. The measured object needs to be positioned in certain positions relative to the measuring device, which requires complex and time-consuming actions. Each change of position also requires a certain time, which can cause increased costs in control and production processes. In order to reduce these costs, several options for positioning a measured object inside the working area of a CMM using an industrial robot have been considered.

In a modern industrial environment the majority of robots are automated systems controlled by computers. Industrial robots have one or more robotic arms, control devices with memory, and sometimes use sensors for data acquisition. They usually support the manufacturing process by positioning objects during machining or welding, transportation, various technological operations, automatic assembly, etc. They are also sometimes used for pre-process, process and post-process control. Industrial robots are widely used in processes which require high quality and productivity.

Research along these lines has been conducted by Santolaria and Aguilar [1]. They conducted a survey about the development of kinematic modelling of robotic manipulators and articulated arm coordinate measuring machines (AACMM), taking into consideration the influences of the chosen model on procedure parameters. Their optimization algorithm included the terms linked to the accuracy and repeatability of the procedure presented. The algorithm follows the simple optimization scheme of data obtained by investigation in several spheres of objects placed at various positions within the working area of both systems.

The majority of research mostly concentrates on differences in the influences on measurement uncertainty in Coordinate Measuring Machines. Lawford [2] observed dysfunctional CMS software with unknown measurement uncertainty and compared its influence with measurement results. He also investigated the prescribed testing algorithms and the checking of industrial software, testing by comparison with given algorithms. This yielded solutions to similar problems.

Fang, Sung and Lui [3] observed the influence of measurement uncertainty from CMM calibration and temperature in the working environment. If the measurement uncertainty in CMM calibration is reduced, this will also reduce the measurement uncertainty of the machine itself. They also pointed out the importance of temperature balance in the working environment before the measurement is performed, i.e., the temperature should be controlled in order to fit CMM working specifications. CMM uncertainty can be reduced using highly precise instruments such as a laser interferometer.

One of the first papers about CMM error compensation was presented by Zhang et al. [4]. They described the error compensation on bridge-type CMM, which resulted in an improvement of accuracy by a factor of 10. They also presented the correction of vector of equally distributed points in the measured volume.

Software error compensation has been reported by a number of authors. Ferreira and Liu [5], for example, developed the analytical model for geometric errors of the machining assembly; Duffie and Yang [6], meanwhile, invented a method to generate the kinematic error function from volumetric measurement error using a vectorial approach.

Robot-CMM integration was performed for the first time by the Mitutoyo Company [7]. They developed a software module in order to adjust the actions of CMM and robotic handling machines used for manipulation of measured parts. Mitutoyo has released the source code in the hope that third party software vendors will be able to use it as a basis to develop products. To the authors' knowledge, at the time of writing no achievements have been made in this direction.

Hansen et al. [8] estimated measurement uncertainty of a hybrid system consisting of an Atomic force microscope attached to a coordinate measuring machine, using linear combination of these two components. Although Hansen et al. also combined two devices, their approach nevertheless differs from ours: we use one system to position the measured object, while they used a two-component system to perform the measurement.

Aggogeri et al. [9] used simulation and planned experimentation to assess the measurement uncertainty of CMMs. They identified and analysed five influence factors, and showed that simulation can successfully be used to estimate CMM uncertainty.

Weckenmann et al. [10] investigated how measurement strategy affects the uncertainty of CMM results. They defined the measuring strategy in relation to "operator influence", which has been neglected in other research. This work showed that measuring strategy influences CMM uncertainty, and that scanning capabilities of modern CMMs, using significantly a larger number of touch points, overcome this influence.

Wilhelm et al. [11] also investigated the influence of measurement strategy, which they defined as the "task specific uncertainty". They also showed that virtual CMM, using Monte Carlo simulation, can be used to estimate uncertainty. Nevertheless, although these authors mentioned that part fixture influences uncertainty, they did not analyse this thoroughly.

Feng et al. [12] applied the factorial design of experiments (DOE) to examine measurement uncertainty. They also studied the effect of five factors and their interaction, and showed that there is statistically significant interaction between speed and probe ratio. They also showed that uncertainty is minimized when speed is highest, stylus length is shortest, probe ratio is largest, and the number of pitch points is largest.

Piratelli-Filho and Giacomo [13] proposed an approach based on a performance test using a ball bar gauge and a factorial design technique to estimate CMM uncertainty. They investigated the effect of length, position, and orientation in work volume on CMM measurement errors. The analysis of variance results showed a strong interaction between the orientation and measured length.

Unlike the mentioned studies, the goal of the research presented here was to assess whether robots can be used to position the measured object in complex measuring systems, using measurement uncertainty analysis and estimating the factors affecting it.

## 2. Problem description

Fig. 1 shows a typical assembly used to clamp the measured object in a CMM. Such a system requires multiple measuring and clamping operations when different features of the measured object are being measured, especially if the geometry is complex and some portions are inaccessible by CMM probes. Such products should be measured in multiple steps.

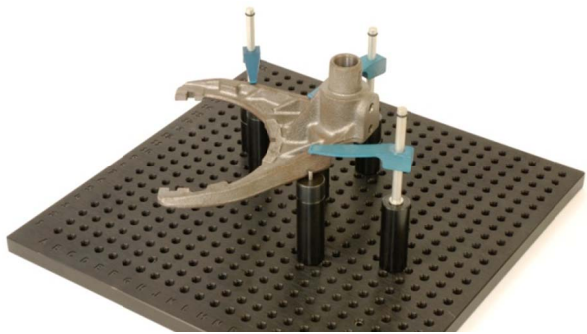

**Figure 1.** Typical fixturing assembly for CMM measurements

The major time-consuming task in coordinate measurement of complex geometry with CMMs is to choose the position of the measured object and the optimal combination of probe stylii, which often prevent the touch probes from coming into contact with the measured object. The obscured surfaces are difficult to reach without repositioning the measured object. Every repositioning of the measured object introduces new error sources in the measurement. Our idea is to use the industrial robot to manipulate the measured object, i.e., to position it automatically.

In order to test this possibility, the measurements were conducted with three different systems:

1. Measuring fixed object with CMM only;
2. Measuring with complex-CMM-robot system, with fixed mass of measured object;
3. Measuring with complex CMM-robot system, with variable mass of measured object.

It is common to use the measurement uncertainty to quantify the quality of the measurement process. Therefore, we decided to estimate it for these three systems, and to compare them with the uncertainty of the robot and the CMM as stated by the manufacturers.

### 2.1 Hypotheses

In order to prove whether it is possible to use a robot, with structurally and technologically limited options, in procedures of precise measurement with a CMM, at the beginning of the research the following three hypotheses were set:

- **Hypothesis 1:** It is possible to use the first generation robot with five degrees of freedom for positioning measured object on CMM.
- **Hypothesis 2:** Uncertainty of measurement for complex CMM-robot measuring system is within limits of allowed uncertainty of measurement of CMM.
- **Hypothesis 3:** Uncertainty of measurement for complex CMM-robot measuring system depends on measured object mass.

These hypotheses were tested by analysing the measurement uncertainty of the complex CMM-robot measuring system, comparing them with limits of allowed measuring uncertainty of CMM (hypotheses 1 & 2) and by varying mass of measured object to analyse the measurement uncertainty when the mass is changed (hypothesis 3). Other influences, such as geometrical error, deformation, thermal error, measuring strategy, probe movement speed during measurement, measuring dynamics, workpiece properties, vibrations, temperature change, etc., were not considered in this research.

### 2.2 Objective

The primary objective of this research was to open new possibilities in this field and to encourage improvements in the capabilities of CMM machines in terms of shortening the procedure and reducing the measurement cycle duration. We tried to point out the possibility of combining different structural solutions on modern and precise equipment in order to achieve fast, reliable and precise measurement, and thus improve technical and technological capabilities in industry, production and research areas where CMMs are commonly used.

## 3. Experiment description

### 3.1 Equipment used

The coordinate measuring machine Zeiss Contura G2 700 Aktiv with tactile probing system was used in this research (measurement range: 700x1000x600 mm, measurement uncertainty according to ISO 10360-2: MPE_E = (1.8+L/300 μm, MPE_P = 1.8 μm).

The positioning and rotating of the measured object was performed manually, inside the CMM's workspace, which tended to take a considerably long time. In order to shorten this time, to reduce the positioning error and to minimize other errors, an educational robot with five degrees of freedom was used: Robot RV-2AJ, manufactured by Mitsubishi Electric–Melfa robots, Japan. The measurement uncertainty of this robot is not stated by the manufacturer; the only comparable parameter is repeatability, stated to be ±0.04 mm. The robot is a stationary robotic system, with programmed motion path and automatic determination of the target.

## 3.2 Conditions

All measurements were performed with the conditions and capacities available at the laboratory at the University of Zenica. The temperature during the experiment was 21°C. The workpiece and CMM measuring elements were cleaned prior to measurement in order to remove possible contaminants. There were no other machines in the vicinity of the CMM; nor were there any other vibration sources (except the CMM's and robot's own vibrations). Prior to measurement, the calibration of the measuring tools and measuring system was performed using 25 mm ceramic reference spheres manufactured by Zeiss, using the calibration procedure defined by CMM software Calypso.

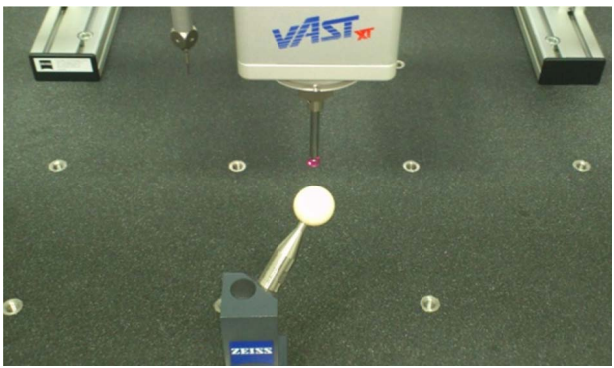

**Figure 2.** Calibration with ceramic reference sphere.

## 3.3 Measured object

The measured object was selected to have geometrical features typically found in coordinate measurements: planes, cones, and cylinders. The material of the measured object was not of great importance, since temperature deviations were negligible (laboratory conditions), and typical measurement force (200 mN) did not deform the object. The object was made of PVC and the surface was metalized, reducing surface roughness to a minimum. Fig. 3 shows the four features measured.

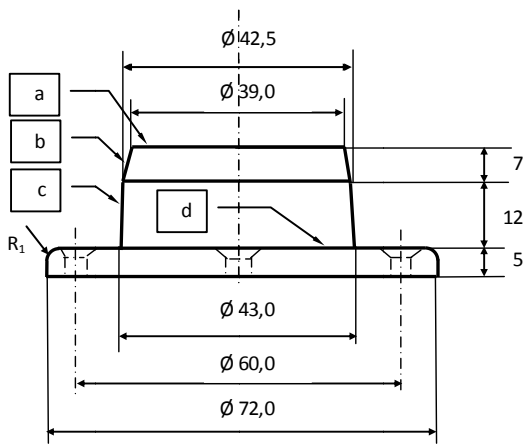

**Figure 3.** Measured object with defined geometrical features (a - top plane of cone, b - conical portion with larger angle, c - conical portion with smaller angle and d - top plane of cylinder)

## 3.4 Positioning

The robot's position compared to CMM was limited by the robot's arm-reach limit or its workspace. Accordingly, the robot was positioned and fixed in an optimal position. This position was defined by relative angular rotation in the arm's joint, and the position and orientation of the grippers in the space, ensuring correct performance of the given assignment. The robot was attached to the CMM's granite table using Z-shaped profile elements with firm screw connections.

Control of the robot was semiautomatic. The control program (direct programming) for piece positioning was followed with manual launching, starting the program for each single measuring phase. After the robot trapped the measured object with the pneumatic gripper, it was then moved into a position enabling measurement with a single stylus system, making all geometrical features easily accessible by all touch probes assembled in the stylus system.

## 3.5 Geometrical features

Dimensional measurements were repeated a certain number of times on previously described surfaces defining different workpiece geometries. For each particular surface, the dimensions were measured 25 times, under the same conditions, in order to compensate random errors. The number of measurements (the size of the sample) was determined according to the significance level of the test $\alpha = 0.01$ and the probability of failing to detect a shift of one standard deviation $\beta = 0.01$ for a two-sided test, assuming normal distribution and known standard deviation [14].

The planar features were measured by sets of 250 points distributed circularly, and conical features were measured by measuring two circles, each consisting of 250 points, at distances of 1 mm (cone "b") and 3 mm (cone "c") from the edges, in order to avoid filleted edges. The measurement results were used to estimate the measurement uncertainty, as a measure of validity of the results and confirmation of hypotheses 1 and 2.

For hypothesis 3, the procedure was identical, but with increased mass of measured object. The first measuring cycle was performed on a CMM with the measuring object fixed on the CMM's granite table. The second and third cycles were measured by the complex CMM-robot system.

Figs. 4 and 5 show the planes and cones used to define the dimensions to be measured.

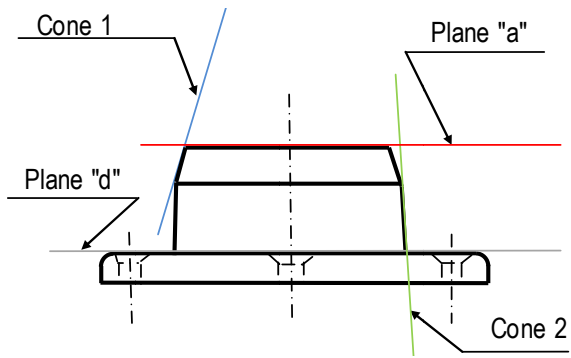

**Figure 4.** Geometric features measured by CMM.

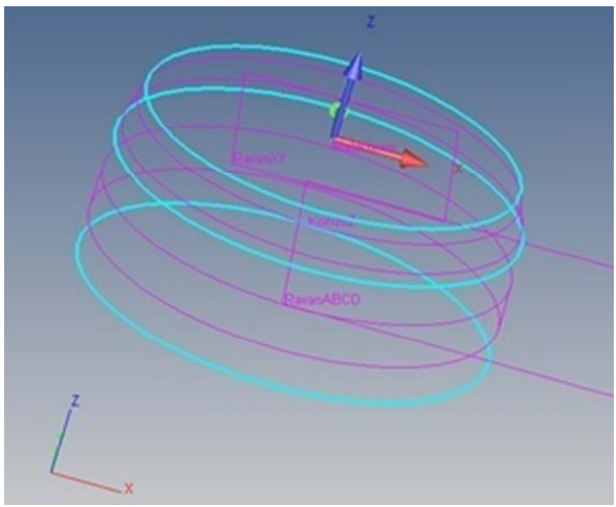

**Figure 5.** Measured features in CMM software.

The measured dimensions were defined as follows:

- Diameter $d_1$ is the intersection of plane "a" and cone 1
- Diameter $d_2$ is the intersection of cone 1 and cone 2
- Diameter $d_3$ is the intersection of plane "d" and cone 2
- Height H is the distance between planes "a" and "d".

## 4. Experiment

In the first measuring cycle, the measured object was positioned and fixed to the CMM's measuring table, and in the second cycle the position of the measured object was defined by the robot's arm position (i.e., auxiliary elements in the robot's arm were holding the measured object) inside the CMM's coordinate space. Between every single measurement in this measuring cycle came a phase of the robot's arm movement from one position to another and back again. Coordinates of the robot's arm in both positions were defined by the robot's off-line programming in such a way that the robot could be positioned manually and that position memorized. After this, the robot's operating speed was defined. Since the robot repeated this operation for each measurement, the robot's repeatability was ±0.04 mm. Figs. 6 and 7 show the first and the second robot arm positions, respectively.

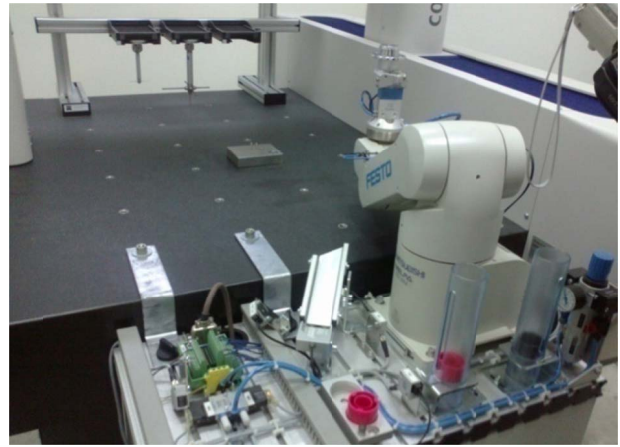

**Figure 6.** Robot arm position 1.

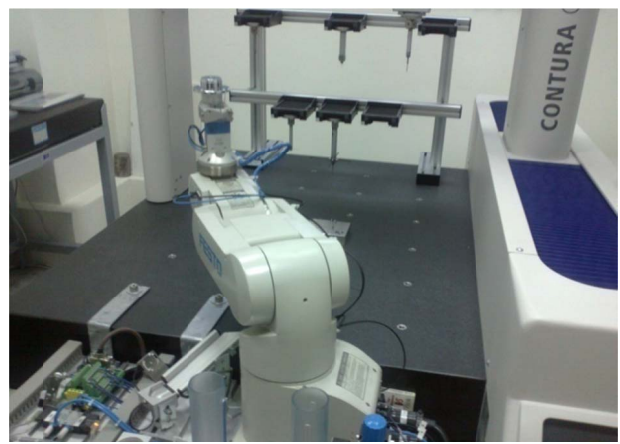

**Figure 7.** Robot arm position 2.

Since the measured object's mass was constant during the first two measuring cycles, the mass of the measured object was increased by adding mass m = 600 g (Fig. 8). The third measuring cycle was conducted with increased mass and the results were compared with the first cycle.

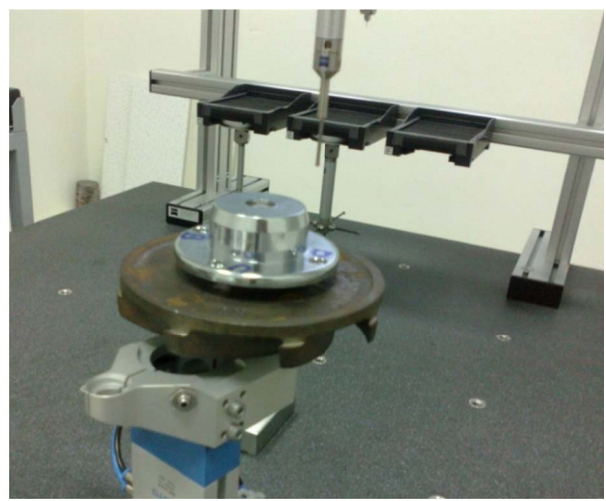

**Figure 8.** Measuring object with additional mass.

## 5. Measurement results

Measurement results are shown in Table 1.

| Measure-ment No. | Measured values (mm) | | | |
|---|---|---|---|---|
| | Diameter $d_1$ | Diameter $d_2$ | Diameter $d_3$ | Height H |
| 1. | 39.0485 | 42.5430 | 42.9402 | 18.8070 |
| 2. | 39.0482 | 42.5432 | 42.9400 | 18.8065 |
| 3. | 39.0484 | 42.5430 | 42.9401 | 18.8067 |
| 4. | 39.0483 | 42.5431 | 42.9401 | 18.8068 |
| 5. | 39.0485 | 42.5433 | 42.9404 | 18.8072 |
| 6. | 39.0484 | 42.5432 | 42.9405 | 18.8070 |
| 7. | 39.0484 | 42.5434 | 42.9407 | 18.8071 |
| 8. | 39.0486 | 42.5436 | 42.9408 | 18.8072 |
| 9. | 39.0486 | 42.5436 | 42.9410 | 18.8073 |
| ... | ... | ... | ... | ... |
| 23. | 39.0481 | 42.5439 | 42.9405 | 18.8071 |
| 24. | 39.0481 | 42.5440 | 42.9404 | 18.8081 |
| 25. | 39.0487 | 42.5437 | 42.9409 | 18.8074 |
| Mean value $x_{1m}$ | 39.0484 | 42.5436 | 42.9407 | 18.8074 |
| Standard deviation | 0.00026 | 0.00038 | 0.00043 | 0.00050 |
| Max | 39.0487 | 42.5444 | 42.9415 | 18.8082 |
| Min | 39.0478 | 42.5430 | 42.9400 | 18.8065 |
| Absolute range $E_1$ | 0.0010 | 0.0014 | 0.0015 | 0.0017 |

**Table 1.** Results of first measuring cycle - measuring object fixed on CMM's measurement table

*5.1 Measurement uncertainty*

The declared measurement uncertainty of the CMM used in this experiment is 1.80 μm. In all three cases there is Type A standard measurement uncertainty, which equals standard deviation times coverage factor 2. The standard measurement uncertainties of three measurement cycles are shown in Table 2 and Fig. 9.

| Measured value | Standard measurement uncertainty (μm) | | |
|---|---|---|---|
| | Case 1 only CMM | Case 2 CMM-robot | Case 3 CMM-robot with added mass |
| Diameter $d_1$ | 0.53 | 12.33 | 3.86 |
| Diameter $d_2$ | 0.76 | 12.98 | 4.69 |
| Diameter $d_3$ | 0.86 | 10.92 | 3.95 |
| Height H | 1.00 | 9.72 | 4.17 |

**Table 2.** Standard measurement uncertainty in three measuring cycles: 1. measuring only on CMM; 2. measuring on CMM-robot system; 3. measuring on CMM-robot, with additional mass.

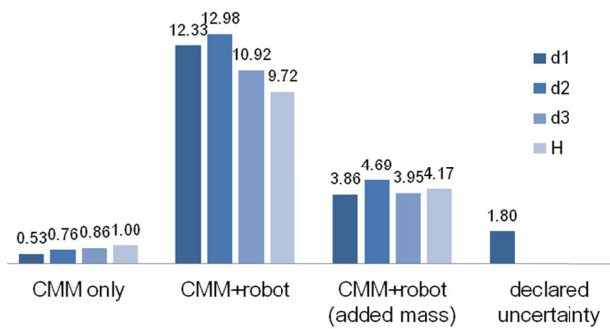

**Figure 9.** Comparison of measurement uncertainties (μm).

We can conclude that all measurements in the first case expressed lower uncertainty than stated on the CMM's calibration certificate. The measurement results in the second case show that measurement uncertainty of the system CMM-robot is significantly larger than the declared uncertainty of the CMM. In the third cycle, where the CMM-robot system was used to measure the object with increased mass, the measurement uncertainties are larger than the declared uncertainty of the CMM, but still lower than those in the second case. This means that increased mass of the measured object slightly reduced uncertainty. This phenomenon could be explained by the increased inertia of the measured object, which stabilizes the system and leads to more accurate results.

*5.2 Statistical analysis*

The first step in statistical analysis was to question the normality of distribution of the measurement results. Kurtosis of all results was between -1.40 and 0.54, and the skew ranged between -0.99 and 0.72. For 25 samples, the standard error of the skew is 0.49 and standard error of the kurtosis is 0.98; therefore, both skew and kurtosis are lower than twice the standard error, and we can assume normal distribution of measured data.

The histograms of distributions for measured values of diameter $d_1$ (Figs. 10 and 11) illustrate the normality of distribution. The distributions of most other measured values have a similar shape.

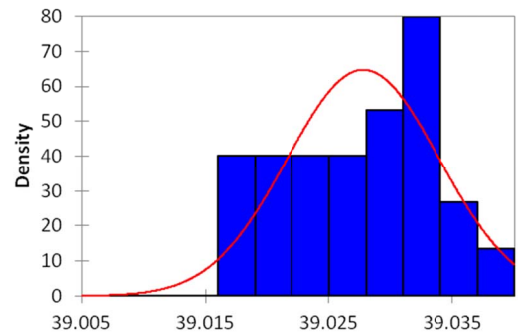

**Figure 10.** Histogram of distribution of diameter $d_1$ (Case 2: CMM-robot) with fitting normal distribution.

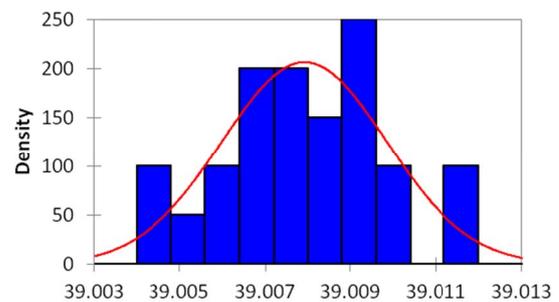

**Figure 11.** Histogram of distribution of diameter $d_1$ (Case 3: CMM-robot + added mass).

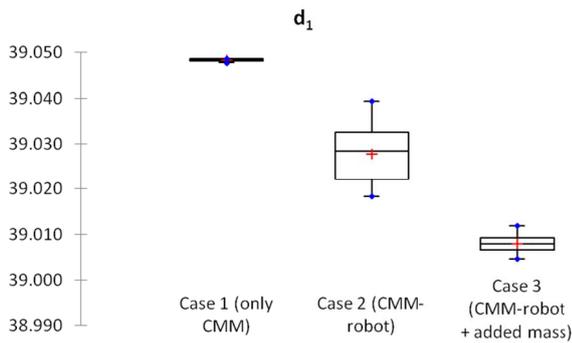

**Figure 12.** Boxplots of diameter $d_1$ reveal no outliers.

Kolmogorov-Smirnov test was used to check the difference in the means of results obtained in the first (measurements performed on fixed measured object using only CMM) and in the second measurement cycle (measurement obtained by the system robot-CMM). The results for all four measured geometrical values (diameters $d_1$ to $d_3$ and height H) are shown in Table 3.

|  | $d_1$ | $d_2$ | $d_3$ | H |
|---|---|---|---|---|
| Mean CMM only | 39.0484 mm | 42.5436 mm | 42.9408 mm | 18.8075 mm |
| Mean CMM-robot | 39.0278 mm | 42.5194 mm | 42.9214 mm | 18.8238 mm |
| Variance CMM only | 6.97E-08 mm | 1.46E-07 mm | 1.86E-07 mm | 2.48E-07 mm |
| Variance CMM-robot | 3.80E-05 mm | 4.21E-05 mm | 2.98E-05 mm | 2.36E-05 mm |
| D CMM only | 0.166 | 0.107 | 0.091 | 0.129 |
| p-value CMM only | 0.458 | 0.920 | 0.980 | 0.771 |
| D CMM-robot | 0.109 | 0.116 | 0.201 | 0.151 |
| p-value CMM-robot | 0.912 | 0.871 | 0.235 | 0.585 |
| t Stat | 16.7275 | 18.5872 | 17.6842 | -16.7088 |
| P(T<=t) 1-tail | 4.94E-15 | 4.69E-16 | 1.43E-15 | 2.24E-15 |
| t Critical 1-tail | 2.492 ($\alpha<0.01$), 1.711 ($\alpha<0.05$) | | | |
| P(T<=t) 2-tail | 9.89E-15 | 9.38E-16 | 2.87E-15 | 4.48E-15 |
| t Critical 2-tail | 2.797 ($\alpha<0.01$), 2.064 ($\alpha<0.05$) | | | |

**Table 3.** The results of the statistical Kolmogorov-Smirnov test (significance level p < 0.05) and t-Test: Two-Sample Assuming Unequal Variances ($\alpha<0.01$, Hypothesized Mean Difference 0).

As the computed p-value is greater than the significance level 0.05, we cannot reject the null hypothesis H0 (the sample follows a Normal distribution).

Levene's test confirmed that variances in the two observed cases are different. Therefore the Welch's t-Test, Two-Sample Assuming Unequal Variances, was performed in order to check the difference in the means of the results obtained in the first and the second measurement cycles.

The results of the t-Test, shown in Table 3, lead to the conclusion that Hypothesis 2 ("Uncertainty of measurement for complex CMM-robot measuring system is within the limits of allowed uncertainty of measurement of CMM") should be rejected, since P-value for both one-tail and two-tail are significantly lower the than critical value of t-variable for sample size 25, $\alpha$ being either 0.05 or 0.01.

*5.3 Discussion and proposed further research*

Although this example shows that it is possible to use an industrial robot to extend the manipulation capabilities of a coordinate measuring machine, some important aspects should be considered. The experiment performed had some disadvantages, which are summarized below.

Disadvantages that could have affected the accuracy of results included:

1. Conditions of university laboratory
   Better equipment, laboratory completeness and application of highest measuring standards can provide better conditions and thus the quality of measurement results.
2. The bonding CMM and robot
   The combination of the robot with the CMM was achieved as described in this paper because of technical and construction capacity constraints. In order to increase stability and measuring precision of the robot, it is possible to use different designs of bearing table using complex binding elements, which can enable the CMM and robot to bind as a single unit.
3. Vibrations caused by the robot's instability on its bearing table (light construction, high position of robot's gravity centre by z axis, wheels on bearing base, etc.).
   Larger and heavier construction of bearing table and a stronger link to the ground would possibly reduce vibrations during movement of movable elements of the robot or CMM. The configuration where the robot was fixed to the ground, without a physical connection to the CMM granite table, drastically increased the system's stability and reduced vibrations.
4. Limited reach of the robot's arm due to its position relative to the CMM.
   By using a different design of CMM and robot binding, or even a different type of robot, it would be possible to increase the overlapping workspace zones of the robot and the CMM, thus increasing the robot's reach.
5. Mechanical impacts on CMM which occur while shifting position of measuring probe in phase of measuring new geometrical feature (surface).
   The design of the CMM used can cause the appearance of certain mechanical impacts when changing measuring phase. These impacts can cause vibration increase in the robot's arm, particularly when at full stretch.

Future research in this area should be performed with different configuration, with a more robust robot chassis, and with more positions examined. Another improvement would be to synchronize the software for CMM manipulation and the software for robot manipulation, providing real automation of the measurement process.

A deeper and more detailed measurement uncertainty analysis, using both Type A and Type B errors, and taking into consideration correlation of influence factors, should also be performed, in order to give a more general foundation for testing the complex measurement systems.

## 6. Conclusion

The principal idea of this paper was to extend the possibilities for automating the measurement process with coordinate measuring machines. The obstacle most often encountered with CMM measurements are limitations of geometry, requiring more measurement sequences in order to reach difficult places on the measured object. It is possible to perform measurements of such objects, but manual repositioning of the measured object, including redefinition of the local coordinate system, slows down the process. If an industrial robot is used to manipulate the measured object, such a process could be automated. The ultimate goal is to keep the measurement uncertainty within allowable limits The measurements of the dimensions of the measured object were conducted by complex CMM-robot measuring system, with movements performed between each single measurement. These results were compared with the results obtained by measuring the same object fixed in the CMM. The measurement results in these two cases were different; one of the reasons for this could be the slight impacts and vibrations that were obvious during every movement phase between measurements.

Although the obtained measurement results still have great accuracy and precision, they do not meet the criteria of the CMM's prescribed measurement uncertainty.

It can be concluded that it is possible to conduct measurements using complex CMM-robot measuring systems, but the measurement results are dictated by the measurement uncertainty of the least accurate component of the system, which in this case was the industrial robot. Significant differences and deviations in measurement results can be confirmed by comparing obtained measurement results with results measured on an object with a different mass. This confirms the significant influence of variation in the mass of the measured object on the measurement uncertainty of the complex CMM-robot measuring system.

It can be argued that it could still be possible to confirm hypothesis 2, assuming the fulfilment of certain conditions such as:

– Different design of CMM and robot combination, which would reduce impacts and vibrations occurring in CMM operation;
– Use of newer and more advanced generations of robots with greater capacity, stiffness, accuracy, repeatability, etc.

## 7. Acknowledgements

This research was supported in part by the Ministry for Education and Science of the Federation of Bosnia and Herzegovina.